\documentclass[letterpaper, 10 pt, conference]{ieeeconf}  %

\IEEEoverridecommandlockouts                              %

\usepackage{cite}
\usepackage{amsmath,amssymb,amsfonts}
\usepackage{algorithmic}
\usepackage{graphicx}
\usepackage{textcomp}
\usepackage{xcolor}
\def\BibTeX{{\rm B\kern-.05em{\sc i\kern-.025em b}\kern-.08em
    T\kern-.1667em\lower.7ex\hbox{E}\kern-.125emX}}

\definecolor{TUMBlue}{RGB}{0,101,189}%
\definecolor{TUMWhite}{RGB}{255,255,255}%
\definecolor{TUMBlack}{RGB}{0,0,0}%
\definecolor{TUMBlue1}{RGB}{0,51,89}%
\definecolor{TUMBlue2}{RGB}{0,82,147}%
\definecolor{TUMGray1}{RGB}{51,51,51}%
\definecolor{TUMGray2}{RGB}{127,127,127}%
\definecolor{TUMGray3}{RGB}{204,204,204}%
\definecolor{TUMBlue3}{RGB}{100,160,200}%
\definecolor{TUMBlue4}{RGB}{152,198,234}%
\definecolor{TUMIvory}{RGB}{218,215,203}%
\definecolor{TUMOrange}{RGB}{227,114,34}%
\definecolor{TUMGreen}{RGB}{162,173,0}%

\usepackage{tikz}
\usepackage{pgfplots}
\usepackage[per-mode=fraction, detect-weight=true]{siunitx}
\usepackage[hidelinks]{hyperref}
\usepackage{subcaption}

\pgfplotsset{
	compat=1.17,
	line-plot/.style={
			width=0.97\columnwidth,
			height=0.625\columnwidth,
			grid=major,
			grid style={dashdotted},
			generic-linestyle/.style={thick}
		},
	box-plot/.style={
			grid=major,
			width=0.85\columnwidth,
			height=0.65\columnwidth,
			grid style={dashdotted},
			generic-linestyle/.style={thick},
			every axis plot/.append style={fill=white,draw=black},
			boxplot/draw direction=y,
			ymin=0,
			x tick label style={
					rotate=60,anchor=east},
			boxplot/box extend=0.65,
			boxplot/.append style={mark=x}
		},
	plot_over_s/.style={
			line-plot,
			xlabel=$s$ in \si{\meter},
			enlarge x limits=0,
			xmin=0,
			y label style={yshift=-0.6em},
			legend columns=2,
			legend style={
					legend pos=south west,
				},
			reference/.style={generic-linestyle, TUMBlack},
			comparison/.style={generic-linestyle, TUMOrange},
		},
	metric_over_ay/.style={
			line-plot,
			xlabel=$a_y$ in \si{\meter\per\square\second},
			enlarge x limits=0.05,
			y label style={yshift=-0.6em},
			legend style={
					legend pos=north west,
				},
			line-1/.style={generic-linestyle, TUMBlack},
			line-2/.style={generic-linestyle, TUMOrange},
			line-3/.style={generic-linestyle, TUMBlue},
			line-4/.style={generic-linestyle, TUMGreen},
			line-5/.style={line-1, dashed},
			line-6/.style={line-2, dashed},
			line-7/.style={line-3, dashed},
			line-8/.style={line-4, dashed},
		},
}

\usetikzlibrary{shapes}
\usetikzlibrary{positioning}
\usetikzlibrary{pgfplots.statistics}

\tikzset{
	flow-chart/.style={
			fc-block/.style={minimum height=1cm, minimum width=3.5cm, draw=TUMBlack, align=center},
			process/.style={fc-block, rectangle},
			process-predef/.style={fc-block, predproc},
			decision/.style={fc-block, diamond, aspect=2},
			flow-fusion/.style={decision, minimum width=1cm, aspect =1},
			input-output/.style={fc-block, trapezium, trapezium left angle = 65,trapezium right angle = 115,trapezium stretches},
			termination/.style={fc-block, rounded rectangle},
			flow/.style={-latex},
			flow-label/.style={fill=white, pos=0.5}
		},
	simulation-architecture/.style = {
			standard-block/.style={
					rectangle,
					rounded corners=.2cm,
					minimum height=1cm,
					minimum width=2cm,
					draw=TUMBlack,
					align=center,
				},
			simulation-block/.style={
					standard-block,
				},
			software-block/.style={
					standard-block,
					draw=TUMBlue
				},
			software-block/.style={
					standard-block,
					white,
					draw=TUMBlue,
					fill=TUMBlue,
				},
			data-block/.style={
					standard-block,
				},
			standard-arrow/.style={-latex},
			standard-arrow-label/.style={fill=none, pos=0.5, align=left}
		}
}

\newcommand{\ownvector}[1]{\mathbf{#1}}

\newcommand{\osmetricmaxlaterror}{d_{\mathrm{max}}}
\newcommand{\osmetriclaterrorsimilarity}[1]{d_{\mathrm{dis,\,}#1}}
\newcommand{\osmetriclaterrorsimilarityrealworld}{\osmetriclaterrorsimilarity{\mathrm{real\,world}}}

\newcommand{\vehiclemodelsymbol}{\mathcal{V}}
\newcommand{\modelbaseline}{\ensuremath{\vehiclemodelsymbol_{\mathrm{base}}}}
\newcommand{\modelmfsimple}{\ensuremath{\vehiclemodelsymbol_{\mathrm{mf\,simple}}}}
\newcommand{\modellineartires}{\ensuremath{\vehiclemodelsymbol_{\mathrm{linear\,tires}}}}
\newcommand{\modelcogzero}{\ensuremath{\vehiclemodelsymbol_{\mathrm{cog\,0}}}}
\newcommand{\modelsingletrackcogzero}{\ensuremath{\vehiclemodelsymbol_{\mathrm{single\text{-}track}}}}
\newcommand{\modelnoactuationdelay}{\ensuremath{\vehiclemodelsymbol_{\mathrm{no\,delay}}}}
\newcommand{\modellessgrip}{\ensuremath{\vehiclemodelsymbol_{\mathrm{less\,grip}}}}
\newcommand{\modellessdownforce}{\ensuremath{\vehiclemodelsymbol_{\mathrm{less\,lift}}}}

\newcommand{\rostwo}{ROS\,2}

\newcommand{\etal}[1]{#1 et al.}

\newcommand{\citetwo}[2]{\cite{#1, #2}}

\newcommand{\Eqref}[1]{(\ref{#1})}

\newcommand{\wheelheave}[1]{z_{#1}}

\newcommand{\zs}[1]{z_{\mathrm{s},#1}}
\newcommand{\dzs}[1]{\dot{z}_{\mathrm{s},#1}}

\newcommand{\fx}[1]{F_{x,#1}}
\newcommand{\fy}[1]{F_{y,#1}}

\newcommand{\fzaero}{F_{z,\mathrm{aero}}}

\newcommand{\myaero}{M_{y,\mathrm{aero}}}

\newcommand{\lf}{l_\mathrm{f}}
\newcommand{\lr}{l_\mathrm{r}}

\newcommand{\axlef}{b_\mathrm{f}}
\newcommand{\axler}{b_\mathrm{r}}

\newcommand{\fc}[1]{F_{\mathrm{c},#1}}

\newcommand{\fs}[1]{F_{\mathrm{s},#1}}
\newcommand{\fd}[1]{F_{\mathrm{d},#1}}
\newcommand{\far}[1]{F_{\mathrm{ar},#1}}
\newcommand{\fa}[1]{F_{\mathrm{a},#1}}

\newcommand{\ks}[1]{k_{\mathrm{s},#1}}
\newcommand{\kd}[1]{k_{\mathrm{d},#1}}
\newcommand{\kar}[1]{k_{\mathrm{ar},#1}}

\newcommand{\laero}{l_{\mathrm{aero}}}
\newcommand{\hrp}{h_{\mathrm{rp}}}
\newcommand{\hpp}{h_{\mathrm{pp}}}

\title{\LARGE \bf
Analyzing the Impact of Simulation Fidelity on the Evaluation of Autonomous Driving Motion Control 
}

\author{Simon Sagmeister$^{1}$, Panagiotis Kounatidis, Sven Goblirsch$^{1}$ and Markus Lienkamp$^{1}$%
\thanks{*This work was funded by the Deutsche Forschungsgemeinschaft (DFG, German Research Foundation) – 469341384; 
and by the "Bundesministerium für Wirtschaft und Klimaschutz" within the research project "ATLAS-L4: Automatisierter Transport zwischen Logistikzentren auf Schnellstraßen im Level 4"}%
\thanks{$^{1}$Technical University of Munich, Germany; School of Engineering \& Design, Department of Mobility Systems Engineering, Institute of Automotive Technology\newline{}
Corresponding author: \href{mailto:simon.sagmeister@tum.de}{simon.sagmeister@tum.de}}
}

\newcommand\copyrighttext{%
	\footnotesize \textcopyright 2024 IEEE.  Personal use of this material is permitted.  Permission from IEEE must be obtained for all other uses, in any current or future media, including reprinting/republishing this material for advertising or promotional purposes, creating new collective works, for resale or redistribution to servers or lists, or reuse of any copyrighted component of this work in other works
}

\newcommand\copyrightnotice{%
	\begin{tikzpicture}[remember picture,overlay]
	\node[anchor=south,yshift=10pt, xshift=0pt] at (current page.south) {\fbox{\parbox{\dimexpr\textwidth-\fboxsep-\fboxrule\relax}{\copyrighttext}}};
	\end{tikzpicture}%
    \vspace{-0.35cm}
}

\begin{document}

\maketitle
\thispagestyle{empty}
\pagestyle{empty}
\copyrightnotice{}
\setcounter{footnote}{1}

\begin{abstract}
    Simulation is crucial in the development of autonomous driving software. In particular, assessing control algorithms requires an accurate vehicle dynamics simulation.
    However, recent publications use models with varying levels of detail. This disparity makes it difficult to compare individual control algorithms.
    Therefore, this paper aims to investigate the influence of the fidelity of vehicle dynamics modeling on the closed-loop behavior of trajectory-following controllers.
    For this purpose, we introduce a comprehensive Autoware-compatible vehicle model. By simplifying this, we derive models with varying fidelity.
    Evaluating over 550 simulation runs allows us to quantify each model's approximation quality compared to real-world data.
    Furthermore, we investigate whether the influence of model simplifications changes with varying margins to the acceleration limit of the vehicle.
    From this, we deduce to which degree a vehicle model can be simplified to evaluate control algorithms depending on the specific application.
    The real-world data used to validate the simulation environment originate from the Indy Autonomous Challenge race at the Autodromo Nazionale di Monza in June 2023. They show the fastest fully autonomous lap of TUM Autonomous Motorsport, with vehicle speeds reaching \SI{267}{\kilo\meter\per\hour} and lateral accelerations of up to  \SI{15}{\meter\per\second\squared}.
\end{abstract}

\section{Introduction}

Controlling a vehicle in highly dynamic situations, such as an object evasion maneuver, is challenging, even for human drivers.
It is even more complex in autonomous driving, as many variables of the environment and the car itself are often unknown.
As a result, many approaches to tackle these challenges in autonomous driving exist \cite{paden2016,betz2022,kato2018}.
Testing and comparing these algorithms is non-trivial since they interact with the vehicle in a closed control loop.
Because of this, the behavior of this closed-loop system depends on both the control algorithm and the controlled vehicle \cite{zhang2024}.

\begin{figure}[!t]
    \centering
    \vspace*{0.2cm}
    \includegraphics[width=\columnwidth]{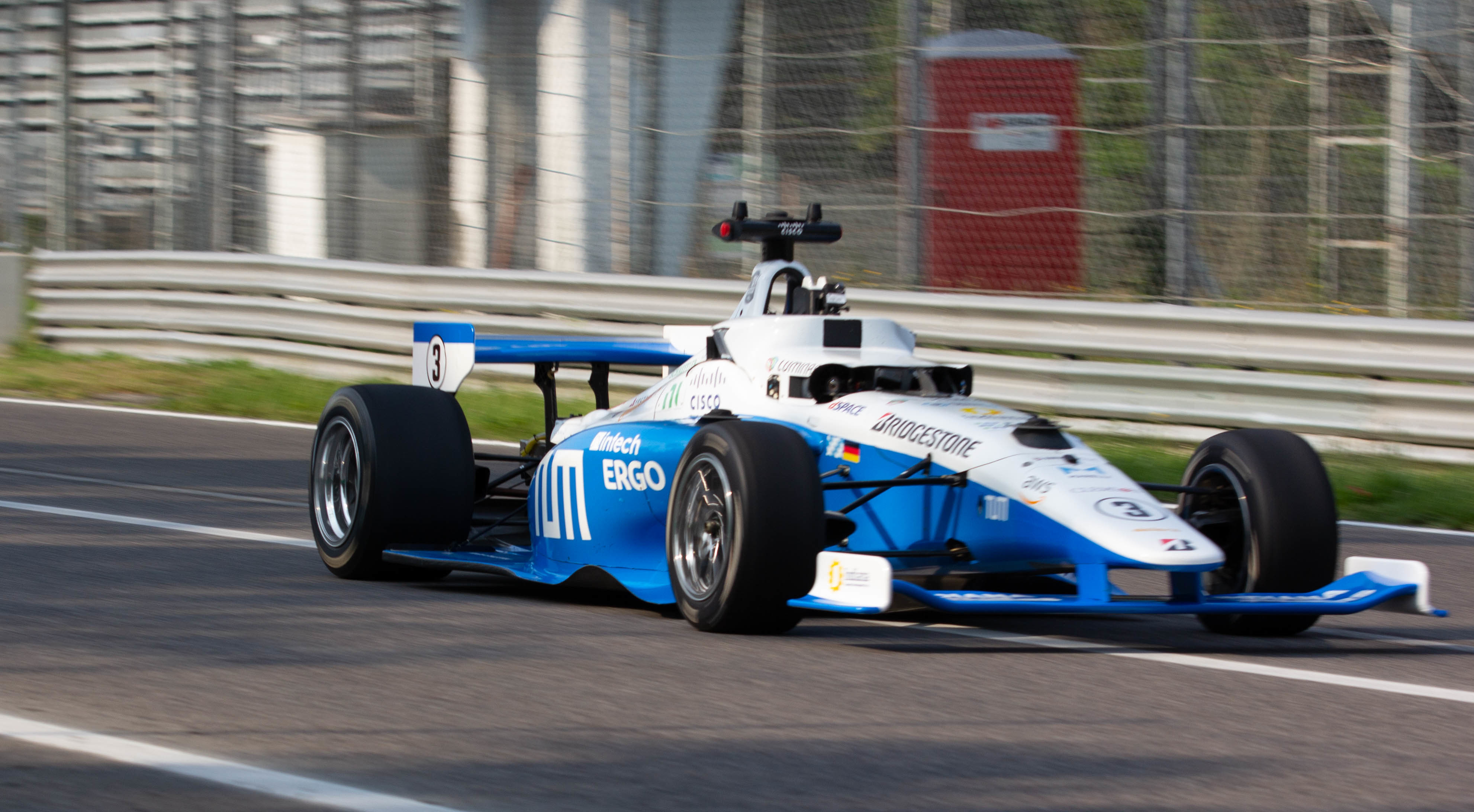}
    \caption{The Dallara AV-21 race car used for data recording on the Autodromo Nazionale di Monza in June 2023.}
    \label{fig:raceboss-image}
\end{figure}

Evaluating these algorithms on a test vehicle additionally imposes specific challenges. Real-world testing is subject to unknown and unpredictable disturbances.
Thus, experiments, even with the same test vehicle, may not be comparable across testing sessions or days. Test time with full-scale autonomous vehicles, such as the one shown in Fig.~\ref{fig:raceboss-image}, is limited.
In addition, operating test vehicles close to their dynamic limits brings the risk of expensive and time-consuming crashes. Consequently, evaluating the performance of these algorithms in simulation is inevitable \cite{hermansdorfer2020a}.

Especially near the limits of handling, modeling a vehicle's dynamic behavior in simulation is challenging \cite{betz2022}. Further, the parametrization of complex models is often impractical due to a lack of data \cite{ning2023a}. Because of this, vehicle models with varying degrees of detail are used \cite{zhang2024}. Since the vehicle model itself influences the behavior of the closed control loop, this severely harms the comparability of different algorithms.

However, the influence of the model accuracy on the closed control loop has not yet been quantified for autonomous driving motion control.
Therefore, this paper comprises the following contributions:
\begin{itemize}
    \item We investigate the closed-loop influence of various simplifications in vehicle modeling, commonly found in literature, against a baseline multibody vehicle model as well as real-world data.
    \item We assess the effect of the simplifications mentioned above for different acceleration reserves concerning the vehicle's dynamic limits. From this, we derive which model fidelity is required depending on the dynamic excitation of the scenario.
\end{itemize}
Additionally, comparing different algorithms can be simplified using a common vehicle dynamics simulation. For this reason, we provide the vehicle model presented in this research as open-source software at \url{https://github.com/TUMFTM/Open-Car-Dynamics}\footnotemark{}.
To maximize compatibility, we implemented the model in C++, fully compatible with the interfaces defined by Autoware \cite{theautowarefoundation}.

\section{Related Work}

Vehicle modeling complexity differs among control-related publications.
This ranges from simple single-track models \cite{hess2013,calzolari2017, rokonuzzaman2021, zarrouki2023} up to only commercially available multibody models \cite{raji2023a, novi2020, zang2022}.
\etal{Zhang} \cite{zhang2024} present a comprehensive overview of the various vehicle dynamics models used for simulation in autonomous racing.

Choosing the best vehicle dynamics model depends on the use case since every model marks a tradeoff between modeling accuracy and effort needed for implementation and parametrization \cite{zhang2024}.
Especially the correct parametrization of complex vehicle models is an ongoing field of research \cite{seong2023,farroni2016,widner2022}.
As a result, multiple publications utilize machine learning algorithms to improve the balance between modeling accuracy and parametrization effort.
For this, the literature presents a variety of methods such as neural networks \citetwo{spielberg2019}{spielberg2022},
physics-informed neural networks \citetwo{kim2022}{chrosniak2023}, bayesian approaches \citetwo{ning2023a}{wischnewski2020a} or reinforcement learning \citetwo{spielberg2023}{kim2016}.

Given a specific use case, knowledge about the sensitivity of certain dynamic effects helps to evaluate the required vehicle modeling fidelity \cite{kutluay2014}. Such studies primarily exist for the open-loop response of vehicle models \citetwo{schmeiler2020}{danquah2022b}.
Additionally, the literature provides similar investigations for the fidelity of the prediction model in Model Predictive Control (MPC) algorithms \citetwo{raji2023a}{liu2016}.

As previously mentioned, publications related to motion control in autonomous driving do not show a consensus on the required vehicle model fidelity. Despite this, no studies exist to evaluate the influence of model fidelity in a closed-loop motion control application.
Additionally, inferring from open-loop investigations to the closed-loop behavior is infeasible \cite{gevers1999}.

Furthermore, \etal{Raji} \cite{raji2023a} show that switching an MPC's prediction model from a single-track to a more sophisticated model improves its real-world performance. It indicates that a single-track model may not consider all dynamic effects relevant for a closed-loop evaluation of a motion controller. The fact that evaluating control performance using a single-track model is considered state of the art \cite{hess2013,calzolari2017, rokonuzzaman2021, zarrouki2023} emphasizes the importance of analyzing the influence of vehicle dynamics modeling on the behavior of the closed control loop.

In summary, the current state of the art does not provide sufficient information on the influence of vehicle modeling in a closed-loop motion control application.

\section{Methodology}

This section outlines our methodology to generate these insights.
Initially, we introduce the simulation environment used to achieve the abovementioned contributions. Combined with the autonomous driving software of TUM Autonomous Motorsport \cite{betz2023}, this simulation enables comparing closed-loop simulation results with real-world data. We use data from the Indy Autonomous Challenge race on the Autodromo Nazionale di Monza in June 2023. An onboard video of the reference lap analyzed in this publication is available online\footnote{\url{https://youtu.be/_FJZ4D2qu5c?si=jMUkim6KsqfL68HI}}.

A detailed description of the used vehicle model and its simplifications follows. After presenting the metrics used to evaluate the quality of the simulation, a description of the conducted experiments concludes this section.
\subsection{Simulation Architecture}

The autonomous driving software of TUM Autonomous Motorsport composes individual modules to realize the task of autonomous driving \cite{betz2023}.
These individual modules are a dedicated \rostwo\,\cite{ros2} node each and run asynchronously in different threads. \rostwo{} enables the communication between these nodes utilizing a middleware layer. For the simulations shown in this paper, CycloneDDS \cite{eclipsefoundation} was used.

\begin{figure}[!tbh]
    {
        \centering
        \vspace*{0.2cm}
        \includegraphics{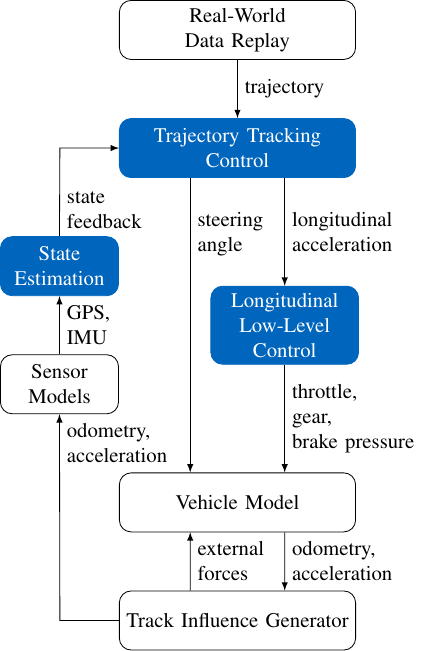}
        \caption{Simulation architecture used in this paper. Blue coloring indicates a component of the in-loop autonomous driving software. Each block corresponds to one \rostwo{} node.}
        \label{fig:sim_architecture}
    }
\end{figure}

Software components dedicated to the driving task rely on the same configuration as during real-world runs, enabling a valid comparison between simulated and recorded data.
Fig.~\ref{fig:sim_architecture} displays these components in blue.
For a description of TUM Autonomous Motorsport's algorithms, which are also open-source available \cite{wischnewski2022}, we refer the reader to the corresponding publications \citetwo{wischnewski2019a}{wischnewski2022c}.
We recently restructured these algorithms to match the interfaces used in the open-source software stack Autoware \cite{theautowarefoundation}. However, since real-world data are only available for the algorithms mentioned above, we cannot conduct the investigations of this work using Autoware, even though the simulation is fully Autoware-compatible.

We replay the target trajectories from the recorded data instead of using the dynamic trajectory planning algorithm to ensure the controller receives the identical trajectory input.
These trajectories resemble a raceline planned under consideration of influences resulting from the three-dimensional track geometry \cite{rowold2023}.

The provisions described above ensure that the in-loop autonomous driving software behaves as during data recording. Consequently, this enables the evaluation of simulation quality by comparing the system's closed-loop response in simulation with the real-world data.

The control algorithm used calculates target values for steering and longitudinal acceleration from the given trajectory. However, actuating the AV-21 race car using longitudinal acceleration is impossible \cite{betz2023}. As a result, we require an additional low-level control algorithm that computes throttle, brake pressure, and the desired gear. Fig.~\ref{fig:sim_architecture} shows that these low-level actuation commands are sent to the vehicle together with the original steering target.
The vehicle model provides complete information about its dynamic state in response to these actuation commands.

Even though banking and slope angles on the Monza track are small, real-world approximation improves when correctly accounting for effects resulting from the three-dimensional track geometry.
We use a distinct module to consider these influences since all vehicle models in this research assume a flat two-dimensional road geometry.
It provides the vehicle model with external forces resulting from the track geometry and the vehicle's pose and velocity.
Furthermore, the track influence generator transforms the two-dimensional output of the vehicle model into the correct three-dimensional representation.

Sensor models calculate the corresponding virtual sensor signals from the three-dimensional vehicle state. The state estimation closes the control loop by providing an estimate of the vehicle's dynamic state to the trajectory tracking algorithm.

\subsection{Vehicle Model}

We introduce a self-developed multibody model as the baseline for the simulations of this work.
This self-developed model allows us to simplify and adapt the model quickly for specific experiments.
The starting point for development is the multibody model of the CommonRoad Vehicle Models \citetwo{althoff2020}{althoff2017a}. However, we adapted the model for our application, initially in \cite{kounatidis2023}, but primarily throughout this work.
The authors aimed to consider as many effects as possible since anticipating the impact of specific dynamic effects on the closed-loop simulation is impossible beforehand.
The following describes the changes and improvements compared to the original model.

We extend the tire model by the previously neglected scaling factors to simplify fitting the model's behavior to real-world data. The formulation corresponds to the 2006 version of Pacejka's magic formula while neglecting tire turn slip \cite[Eqns.~4.E1\,--\,4.E30\,\&\,4.E50\,--\,4.E67]{pacejka2006}. Further, we model tire delay according to \cite{pacejka2006}.

The original model assumes a single unsprung mass for the front and rear axle each \cite{althoff2020}. We adapt this to consider the mass of each tire individually since the real-world race car has an independent lightweight suspension for all four tires. Further, the developed model respects squat and lift forces originating from the suspension geometry \cite{allen1992}.

We model the coupling between the vehicle body and the wheels using a composite suspension force $\fc{i}$.
The suspension force acts vertically, directly above the center of each tire's contact patch.
It consists of the suspension spring $\fs{i}$, damper $\fd{i}$, anti-roll bar $\far{i}$ and axle $\fa{i}$ forces, with the index $i$ indicating the respective wheel.
\begin{equation}
    \fc{i} = \fs{i} + \fd{i} + \far{i} + \fa{i}
\end{equation}
$\fs{i}$ and $\fd{i}$ result from the suspension travel $\zs{i}$ and the corresponding stiffness coefficients $\ks{i}$ and $\kd{i}$ \cite[Eqns.~A23a-A26b]{allen1992}.
\begin{align}
    \fs{i} & = \zs{i}\ks{i}  \\
    \fd{i} & = \dzs{i}\kd{i}
\end{align}

However, since the model has an independent suspension for each wheel, we adapt the calculation of the suspension travel $\zs{i}$ accordingly.
The anti-roll bar force $\far{i}$ is proportional to the difference in vertical travel between left and right.
\begin{align}
    \far{\mathrm{fl}} & = - \far{\mathrm{fr}} = (\wheelheave{\mathrm{fr}}-\wheelheave{\mathrm{fl}})\kar{\mathrm{f}} \\
    \far{\mathrm{rl}} & = - \far{\mathrm{rr}} = (\wheelheave{\mathrm{rr}}-\wheelheave{\mathrm{rl}})\kar{\mathrm{r}}
\end{align}
We neglect suspension-travel-induced changes in geometry since we assume small pitch and roll angles for the modeled highly stiff race car.
With this, the model calculates the squat and lift forces resulting from the geometry of the axle using the following equations \cite[eq.~A32-A35]{allen1992}.
\begin{align}
    \fa{\mathrm{fl}} & = 2\,\fy{\mathrm{fl}}\frac{\hrp}{\axlef} + \fx{\mathrm{fl}}\frac{\hpp}{\lf}   \\
    \fa{\mathrm{fr}} & = - 2\,\fy{\mathrm{fr}}\frac{\hrp}{\axlef} + \fx{\mathrm{fr}}\frac{\hpp}{\lf} \\
    \fa{\mathrm{rl}} & = 2\,\fy{\mathrm{rl}}\frac{\hrp}{\axler} - \fx{\mathrm{rl}}\frac{\hpp}{\lr}   \\
    \fa{\mathrm{rr}} & = - 2\,\fy{\mathrm{rr}}\frac{\hrp}{\axler} - \fx{\mathrm{rr}}\frac{\hpp}{\lr}
\end{align}
Thereby, $\fx{i}$ and $\fy{i}$ are the longitudinal and lateral tire forces in the vehicle body axis system with $\hrp$ and $\hpp$ naming the roll and pitch pivot point height above ground. Since we model independent suspensions, different pitch pivot point heights are considered during acceleration and deceleration \cite{allen1992}.
We model the tire as a spring to determine the vertical force in its contact patch depending on the rim's distance to the road.

To reproduce the aerodynamic properties of the test vehicle, we model forces for drag and lift \cite{gillespie2021}. We also consider an additional pitch moment $\myaero$, which allows for adjusting the aerodynamic balance between front and rear. It is the product of the lift force $\fzaero$ and its distance from the center of gravity $\laero$.
\begin{equation}
    \myaero  = \fzaero\,\laero
\end{equation}
For integration into the simulation environment, the inputs for the vehicle dynamics model have to be adjusted. The CommonRoad multibody model \cite{althoff2020} can be controlled via steering angle speed and longitudinal acceleration. However, as can be seen in Fig.~\ref{fig:sim_architecture}, the autonomous software only provides a target steering angle as well as throttle, gear, and brake pressure for longitudinal actuation.

For this, we model the car's steering actuator by combining time delay and a first-order transfer function.
Accepting the above-mentioned low-level control inputs requires a powertrain model. The implemented powertrain considers the engine map, actuator latencies, and the drivetrain topology of the race car (rear-wheel drive, spool-type differential). It provides the rotational speeds of each wheel as input to the vehicle dynamics model.

Combining all the previously mentioned submodels results in a single state-space model comprising 35 state variables and 14 degrees of freedom.
Due to the scope of this work, we cannot provide a more comprehensive description of the model.
Instead, we refer the reader to the documentation of the published open-source software.
The following equation shows the state equation mentioned above.
\begin{equation}
    \dot{\ownvector{x}} =\ownvector{f}\left(\ownvector{x},\,\ownvector{u}\right)
\end{equation}

We use the Dormand Prince 45 (also known as ode45) integration scheme \cite{dormand1980} to solve the resulting differential equation. Disabling the solver's step size adaptivity enables real-time execution.

On an Intel Core i7-11850H CPU, the abovementioned solver takes a mean of around \SI{95}{\micro\second} per integration step. By measuring code required for integration into the simulation environment, for example, logging debug signals, the execution time increases to around \SI{150}{\micro\second} per step.
The chosen step size for the simulations in this work is with \SI{800}{\micro\second} over five times longer than the mean execution time, enabling reliable real-time simulation.

We used the provided setup data of the test car to parametrize the model. However, certain parameters, such as the tire model or the exact suspension geometry, were unavailable to the authors.
Therefore, it was necessary to empirically fit these parameters using real-world data.
Based on the parameter set of a high-performance road tire (with profile), we adapted the tire model to the measured data using the respective scaling factors in the magic formula.
Since this work aims to evaluate the impact of vehicle modeling on closed-loop control behavior, we opted for a parametrization that matches the closed-loop behavior with the recorded data.
For this, directly fitting the model parameters closed-loop yielded the best results.

\begin{figure}[t]
    \centering
    \vspace*{0.2cm}
    \includegraphics{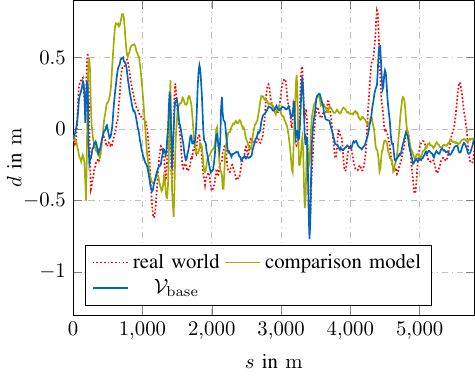}
    \caption{Lateral displacement error $d$ of the tracking control algorithm shown for the recorded real-world data and in simulation. In addition to the model developed during this work \modelbaseline, the plot shows another compatible open-source vehicle model \cite{hermansdorfer2022} for comparison.}
    \label{fig:d_over_s-base_vs_real}
\end{figure}

Fig.~\ref{fig:d_over_s-base_vs_real} shows the achieved modeling accuracy concerning the lateral displacement error of the trajectory tracking controller. Hermansdorfer and Wischnewski \cite{hermansdorfer2022} published another validated vehicle model implementation suitable for modeling a race car.
Therefore, we compare their model to the one developed during this work.

Fig.~\ref{fig:d_over_s-base_vs_real} reveals that our \modelbaseline{} outperforms the already available model of Hermansdorfer and Wischnewski \cite{hermansdorfer2022}.
\modelbaseline{} predicts the real-world lateral displacement error better in almost every turn, with the only exception being in turns six and seven ($s \in [1600, 2300]$).
However, local on-track influences, such as a variation in tire grip, could also cause these deviations in turns six and seven.
Consequently, we choose \modelbaseline{} as the baseline model for the remainder of this work.

\subsection{Model Simplications}

As mentioned in the previous sections, different levels of detail for modeling vehicle dynamics are present in the literature.
Tab. \ref{tab:models} gives an overview of the various models compared within this work.
These reflect common models used in publications to simulate vehicle dynamics in motion control.
In addition, we introduce two models that originate from a slight variation of parameters that are difficult to determine. These two models serve as a comparison and allow for a better interpretation of the influence of certain simplifications in vehicle dynamics modeling.

\begin{table}[!tbh]
    \centering
    \vspace*{0.2cm}
    \caption{The vehicle models compared within this work}
    \begin{tabular}{|c|m{0.6\columnwidth}|}
        \hline
        \textbf{Model}           & \textbf{Difference to \modelbaseline}                                                                                                                                                                                                                                                                    \\ \hline
        \modelmfsimple           & Switching the tire model to the basic version of the magic formula \cite{pacejka2006} with only four parameters for lateral and longitudinal behavior each. We analytically calculate the parameters for the simplified tire model from the original one at the median tire load seen in the simulation. \\ \hline
        \modellineartires        & Switching to a linear tire model. Slip stiffness matches the linear area of the magic formula tire model.                                                                                                                                                                                                \\ \hline
        \modelcogzero            & Assuming the vehicle's center of gravity as well as the pitch and roll pivot points lie in the road plane. As a result, no acceleration-induced tire load transfer occurs.                                                                                                                               \\ \hline
        \modelsingletrackcogzero & Starting from \modelcogzero{}, we change the front and rear track width to zero. This results in the same slip conditions for each axle's left and right tire, resembling a single-track model.                                                                                                          \\ \hline
        \modelnoactuationdelay   & Neglecting the actuation models, leading to undelayed and perfect tracking of the actuators concerning the requested target values.                                                                                                                                                                      \\ \hline
        \modellessgrip           & Reducing grip and slip stiffness of the tires by \SI{10}{\percent}.                                                                                                                                                                                                                                      \\ \hline
        \modellessdownforce      & Adapting the lift coefficient so that downforce decreases by \SI{10}{\percent}.                                                                                                                                                                                                                          \\ \hline
    \end{tabular}
    \label{tab:models}
\end{table}

\subsection{Metrics for Evaluating Simulation Quality}

There is no standardized procedure to validate a vehicle dynamics model.
Instead, Kutluay and Winner \cite{kutluay2014} suggest evaluating its performance based on the values that it attempts to represent.
As we aim to quantify the impact of vehicle modeling on the closed control loop response,
we assess simulation quality by using metrics intended for rating the performance of a motion control algorithm.
In the literature, the mean and the maximum lateral displacement error are the dominant metrics \citetwo{hess2013}{rokonuzzaman2021}.

However, the mean lateral displacement error is unsuited for evaluating this work's motion controller.
The robust tube MPC control algorithm proposed by \etal{Wischnewski} \cite{wischnewski2022c} reoptimizes its target trajectory.
It allows the algorithm to deviate from the planned trajectory as long as its uncertainty-aware prediction stays within a defined driving tube.
However, this "deviation by design" influences the mean lateral displacement error while not correlating with the algorithm's control performance.
We, therefore, choose the maximum lateral error $\osmetricmaxlaterror$ as the control performance metric for the remainder of this work.

\Eqref{eq:max-lat-error} gives the formula for calculating the maximum lateral error $\osmetricmaxlaterror$. Therein, $d$ names the lateral displacement error, while $s$ represents the longitudinal track position expressed in the curvilinear coordinate system of the racetrack's centerline.
\begin{equation}
    \osmetricmaxlaterror = \underset{\mathrm{0 \,\leq\, s\, \leq \,s_{\mathrm{max}}}}{\max}\left|d\left(s\right)\right|
    \label{eq:max-lat-error}
\end{equation}

However, only using the maximum lateral tracking error to assess simulation quality is insufficient since it discards information potentially relevant to other lateral-error-based metrics.
To compensate for this, we introduce an additional metric that compares the similarity between two curves.
\Eqref{eq:lat-error-similarity} describes the disparity $\osmetriclaterrorsimilarity{i}$ to a reference curve $i$, for example, the lateral displacement error from real-world data.
\begin{equation}
    \osmetriclaterrorsimilarity{i} = \frac{1}{s_{\mathrm{max}}} \int_{0}^{s_{\mathrm{max}}}\left(d(\xi) - d_{i}(\xi)\right)^{2}\,\left|d_{i}(\xi)\right|\,\mathrm{d}\xi
    \label{eq:lat-error-similarity}
\end{equation}
It averages the product of the squared difference between the curves and the reference curve's current absolute lateral tracking error.
By scaling with the lateral error reference, we account for the control algorithm's behavior not being meaningful at low lateral errors.

\subsection{Conducted Experiments}

To quantify the influence of vehicle modeling fidelity on control behavior, we investigate the reproducibility and determinism of our simulation environment beforehand.
We use our base simulation model to repeat an identically parametrized simulation 30 times, once with and once without additional sensor noise.
The resulting spread regarding the analyzed metrics provides information about reproducibility.

\begin{figure}[!b]
    \centering
    \vspace*{0.2cm}
    \begin{subfigure}{.45\columnwidth}
        \centering
        \includegraphics{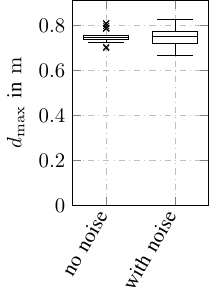}
        \caption{Maximum lateral error.}
        \label{fig:sub1}
    \end{subfigure}%
    \begin{subfigure}{.45\columnwidth}
        \centering
        \includegraphics{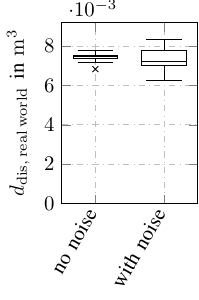}
        \caption{Lateral error disparity}
        \label{fig:sub2}
    \end{subfigure}
    \caption{Reproducibility regarding the analyzed metrics.}
    \label{fig:influence-sensor-noise}
\end{figure}

Fig.~\ref{fig:influence-sensor-noise} shows the distribution of results for both metrics evaluated in this work.
It reveals that the simulation environment is non-deterministic for both evaluation criteria, even when neglecting additional sensor noise.
This lack of determinism has two leading causes.
First, the simulation environment consisting of multiple \rostwo{} nodes is asynchronous and, therefore, influenced by variations in computation time \cite{ros2}.
Second, the solver used to solve the optimal control problem in our motion controller is non-deterministic \cite{stellato2020}.

The first experiment simulates exactly the lap corresponding to the recorded data.
By switching the in-loop vehicle dynamics model and evaluating the metrics above, we assess the simulation quality for each model. We repeat the simulation 30 times for every model to account for the influence of the above-mentioned statistical fluctuations. The autonomous driving motion controller \cite{wischnewski2022c} and its configuration remain unchanged for all the experiments conducted in this work.

As a second experiment, we repeat the first experiment while additionally scaling the lap's velocity profile by a constant factor.
Thereby, the geometric path remains unchanged. The resulting longitudinal acceleration scales linearly with the adapted velocity, while the lateral acceleration scales quadratically.
This way, we can evaluate simulation quality for different models at varying accelerations.
However, no real-world data for these scaled laps are available, so we compare all models to \modelbaseline{}.
We increase acceleration until the control algorithm cannot complete the lap using the baseline model. We repeat each run only five times to reduce the required runs.
To improve reproducibility, we omit imposing sensor noise, which reduces the standard deviation of repeated runs by a factor of $2.8$ while only slightly impacting the mean of the lateral error disparity (Fig.~\ref{fig:influence-sensor-noise}).

\section{Results}

The following section describes the results of the previously described experiments. Before assessing the impact of lateral acceleration, we start by comparing the various models with real-world data.

\subsection{Model Comparison with Real-world Data}

\begin{figure}[!tbh]
    \centering
    \vspace*{0.2cm}
    \includegraphics{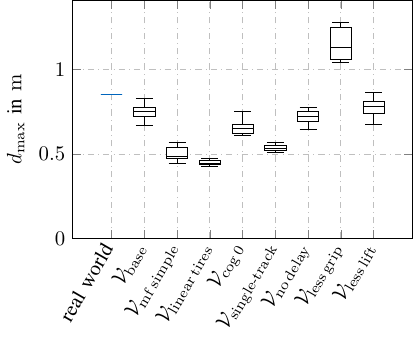}
    \caption{Maximum lateral tracking error for each simulation model during the reference lap. The real-world data shown for comparison represent a single sample.}
    \label{fig:fitting_quality_max_lat_error}
\end{figure}

This subsection evaluates each model's approximation quality by comparing the response of the closed control loop with the recorded data.
Fig.~\ref{fig:fitting_quality_max_lat_error} shows the distribution of the maximum lateral error for each of the analyzed models.
Therein, the best fitting models are \modelbaseline{}, \modellessdownforce{}, and \modelnoactuationdelay{}.
Their mean differs from the real-world value by about \SI{0.1}{\meter}. However, some simulation runs are closer to the recorded sample.
For example, the best fitting \modelbaseline{} run predicts the real-world maximum lateral displacement with a deviation of only \SI{2.7}{\percent}.

In contrast, models with a variation in tire modeling as well as the $\modelsingletrackcogzero$ approximate the real-world data the worst regarding $\osmetricmaxlaterror$. These models are off by a mean of \SIrange{0.3}{0.4}{\meter}, which reflects a percentage error of over \SI{40}{\percent}. Fig.~\ref{fig:fitting_quality_max_lat_error} also shows an increased variance for the model with decreased tire grip. However, it cannot be distinguished if this is due to the modified grip or if the standard deviation generally increases at higher lateral deviations.

Neglecting acceleration-induced dynamic load transfer doubles the offset to the recorded data to a value of approximately \SI{0.2}{\meter}.

Additionally, Fig.~\ref{fig:fitting_quality_max_lat_error} illustrates that all models, except $\modellessgrip$, tend to predict lower lateral errors than the actual car.
However, the maximum displacement error always occurs at the same point on the track. Adapting the parametrization of the baseline model to improve its approximation disproportionately harms the lateral error disparity. A local disturbance not modeled in simulation, such as wind or a dirty road surface, could explain this phenomenon. To determine this with certainty is impossible due to a lack of data.

\begin{figure}[!tb]
    \centering
    \vspace*{0.2cm}
    \includegraphics{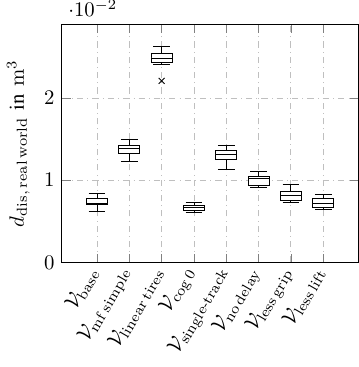}
    \caption{Disparity of the lateral displacement error between each simulation model and the real-world data.}
    \label{fig:fitting_quality_similarity}
\end{figure}

Fig.~\ref{fig:fitting_quality_similarity} depicts a slightly different pattern.
The models portraying a modified parametrization outperform almost all simplified models regarding error disparity.
The only exception is $\modelcogzero{}$.
Considering each model's best runs, it is comparable to the baseline model and achieves an approximately \SI{9}{\percent} lower mean.
Both \modellessdownforce{} and \modellessgrip{} indicate low sensitivity of $\osmetriclaterrorsimilarityrealworld$ against a modified parametrization.
Their mean differs by \num{1} and \SI{11}{\percent} respectively.

The \modelsingletrackcogzero{} is with an average of \SI{0.13}{\meter\cubed} similar to the \modelmfsimple{}. Their mean exceeds over \SI{75}{\percent} relative difference to the \modelbaseline{}.
The vehicle model utilizing a linear tire formulation has a dissimilarity value three times higher than the baseline model and is the worst performing in this test.

\subsection{Influence of Lateral Acceleration}

\begin{figure}[!tb]
    \centering
    \vspace*{0.2cm}
    \includegraphics{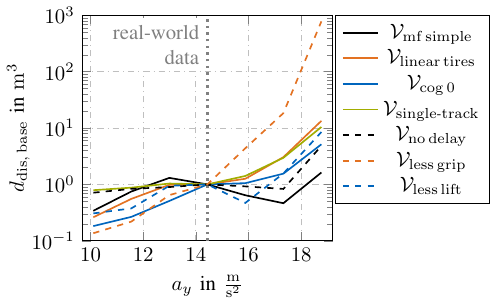}
    \caption{The mean lateral error disparity dependent on the maximum lateral target acceleration.}
    \label{fig:influence-lat-acceleration}
\end{figure}

After examining each model's approximation quality with the unmodified speed profile, we investigate how these findings translate to different lateral accelerations.
For this, Fig.~\ref{fig:influence-lat-acceleration} displays the previously introduced disparity metric over the highest requested lateral acceleration.
The maximum lateral acceleration target on the recorded lap corresponds to approximately \SI{14.4}{\meter\per\second\squared}.

The figure shows a clear trend of increasing difference to the baseline model with rising lateral acceleration for almost all simplified models.
The only exception is \modelmfsimple{}, showing no clear trend regarding a variation in the targeted acceleration. Apart from \modelnoactuationdelay{}, the impact of the various assumptions made in the modeling phase scales exponentially with lateral acceleration towards the acceleration limits. Lowering the dynamic excitation lowers the disparity to the baseline for most models, while \modelmfsimple{}, \modelsingletrackcogzero{}, and \modelnoactuationdelay{} do not profit from decreasing acceleration levels.

Furthermore, Fig.~\ref{fig:influence-lat-acceleration} indicates that most modeling simplifications scale similarly towards the limits of handling.
The \modelnoactuationdelay{} is an exception since a variation in lateral acceleration does not impact the model until up close to the vehicle's dynamic limit.
Additionally, Fig.~\ref{fig:influence-lat-acceleration} shows that a variation in tire grip outweighs all simplifications in vehicle modeling in high acceleration situations by a factor of over~\num{18}. In contrast, \modellessgrip{} is the best-fitting model for low acceleration situations.

\section{Discussion}
\label{sec:conclusion}

We start the conclusion of this work by discussing the results presented in the previous section.
Both the \modelsingletrackcogzero{} model and the \modellineartires{} differ heavily from recorded real-world data, independently of the used metric.
In addition, the influence of the simplifications incorporated in these models increases exponentially with acceleration.
As a result, a comprehensive evaluation of control algorithms is infeasible while using these simplified vehicle models.
Since their impact outweighs all other vehicle model simplifications over a broad range of accelerations, the authors recommend against using such simple models.

Using the basic version of the magic formula tire model also considerably impacts the behavior of the closed control loop.
However, a slight variation in the parametrization of the tire model outweighs the effect of the simplified formulation when approaching the vehicle's dynamic limit.
Additionally, the impact of the simplified magic formula shows no clear dependency on acceleration.
Therefore, assuming tire behavior according to the basic magic formula is a valid fallback if no complete parameter set for the actual tire is available.

Even completely neglecting dynamic load transfer shows less influence than an imprecise tire model.
This indicates that the complex and time-consuming modeling of the suspension geometry and, thus, the wheel load transfer dynamics can be omitted since it is outweighed by the tire parametrization.
However, because its influence scales exponentially with acceleration, we still recommend an abstracted model to consider dynamic wheel load transfer for highly dynamic scenarios.

Neglecting the dynamics of the vehicle's actuators influences simulation similarly to neglecting wheel load transfer for dynamic scenarios.
Furthermore, its impact does not fade away at low accelerations. For these reasons, a closed-loop simulation should incorporate models reflecting the actuator dynamics, especially since
even simple models like the ones described in this work have a positive effect.

The considerable influence of slightly modified tire parameters hints that correct parametrization is more important than considering every dynamic effect in vehicle modeling with absolute detail.

However, we used a highly robust algorithm to generate the above insights.
This controller accounts by design for disturbances or a potential model mismatch.
Using a less sophisticated control approach could further accentuate the disparity between the various models.
Additionally, the above results reflect vehicle modeling for a highly stiff race car.
The influence of certain simplifications may vary slightly for a different vehicle, but we generally expect similar trends.

\section{Conclusion}
In summary, this work showed the influence of certain simplifications in vehicle modeling on the closed-loop behavior of a control algorithm.
We validated our simulation and compared different vehicle models to real-world data.
Further, we examined the impact of vehicle modeling fidelity for different acceleration levels.
This knowledge simplifies the comparison and evaluation of different control algorithms.
We demonstrated that using a detailed and correctly parametrized model enables rating a motion control algorithm in simulation while yielding performance metrics comparable with real-world tests.
Therefore, we provide the model developed for this work as open-source software compatible with Autoware.

\section*{Acknowledgment}
Author contributions: Simon Sagmeister, as the first author,
designed the structure of the article and contributed essentially to the development of the baseline model, simulation, evaluation of results, and the overall contents. Panagiotis Kounatidis and Sven Goblirsch contributed to writing the paper and provided regular feedback during the development phase of the vehicle model.
Markus Lienkamp made an essential contribution to the concept of the research project. He revised the paper critically for important intellectual content. Markus Lienkamp gives final approval for the version to be published and agrees to all aspects of the work. As a guarantor, he accepts responsibility for the overall integrity of the paper.

% Generated by IEEEtran.bst, version: 1.14 (2015/08/26)

\end{document}